\documentclass[a4paper]{article}

\usepackage{mathtools}
\usepackage{setspace} 
\usepackage{dsfont}
\usepackage{amsfonts}
\usepackage{amsmath}
\usepackage{subcaption}
\usepackage{paralist}
\usepackage{times}
\usepackage{latexsym}
\usepackage{graphicx}
\usepackage[T1]{fontenc}
\usepackage{tikz}
\usepackage{tikz-qtree}
\usetikzlibrary{positioning}
\usepackage{url}
\usepackage{pgfplotstable}
\usepackage{titlesec}
\usepackage{color}
\usepackage{lipsum,adjustbox}
\usepackage[font={small}]{caption}
\usepackage{booktabs}
\usepackage{numprint}
\usepackage{bbm}

\usepackage{xr}
\makeatletter
\newcommand{\@BIBLABEL}{\@emptybiblabel}
\newcommand{\@emptybiblabel}[1]{}
\usepackage[hidelinks]{hyperref}

\usepackage[hyperref]{naaclhlt2018}
\def\aclpaperid{***}
\usepackage{naaclhlt2018}
\aclfinalcopy
\graphicspath{{./plots/}}
\newcommand{\com}[1]{}

\newenvironment{myequation*}{

	\begin{equation*}
}{
\end{equation*}

}

\begin{document}

\title{Reference-less Measure of Faithfulness for Grammatical Error Correction}
\author{
  Leshem Choshen\textsuperscript{1} and Omri Abend\textsuperscript{2} \\
  \textsuperscript{1} School of Computer Science and Engineering,
  \textsuperscript{2} Department of Cognitive Sciences \\
  The Hebrew University of Jerusalem \\
  \texttt{leshem.choshen@mail.huji.ac.il, oabend@cs.huji.ac.il}\\
}
\maketitle

\begin{abstract}
  We propose {\sc USim}, a semantic measure for Grammatical Error Correction (GEC)
  that measures the semantic faithfulness of the output to the source,
  thereby complementing existing reference-less measures (RLMs) for measuring the output's grammaticality.
  {\sc USim} operates by comparing the semantic symbolic structure of the source and the correction,
	without relying on manually-curated references.
  Our experiments establish the validity of {\sc USim},
  by showing that (1) semantic annotation can be consistently applied to ungrammatical text; (2)
  valid corrections obtain a high {\sc USim} similarity score to the source; and (3)
  invalid corrections obtain a lower score.\footnote{Our code is available in \url{https://github.com/borgr/USim}.}
  %grammatical reference-less evaluation measure to
  %create a measure balancing between the goal of correcting grammatical mistakes and the goal of
  %conveying the original meaning.
\end{abstract}

%%%%%%%%%%%%%%%%%%%%%%%%%%%%%%%%%%%%%%%%%%%%%%%%%%%%%%%
\section{Introduction}

Evaluation in Monolingual Translation, and particularly in Grammatical Error Correction (GEC) is a challenging
research field, much due to the difficulty in integrating 
different types of rewriting operations into a single measure, and the vast number of valid outputs
\cite{tetreault2008native,madnani2011they,chodorow2012problems,bryant2015far}.
These difficulties have recently motivated
a number of proposals for new, improved reference-based measures (RBMs)  \cite{dahlmeier2012better,felice2015towards,napoles2015ground}.

Nevertheless, the size and heterogeneity of the space of valid outputs per sentence often prohibits
obtaining a reference set that covers this space well, thereby limiting the applicability
of RBMs \cite{bryant2015far}.
To address this we propose a semantic RLM, {\sc USim}, that
operates by measuring the graph distance between the semantic
representations of the source and the output.
Reliable RLMs are appealing both in not relying on references, which are costly to collect, and in avoiding the biases incurred by selecting references that necessarily cannot exhaust the vast space of valid corrections.

Our proposal complements the RLM proposed by 
\newcite{napoles-sakaguchi-tetreault:2016:EMNLP2016}, which uses grammatical error detection techniques to assess the grammaticality of the output, and the work of \newcite{asano2017reference}, who advocate the use of RLMs for fluency, grammaticality and meaning preservation, but state that a meaning preservation measure for GEC is currently lacking.
A similar decomposition of output quality to its adequacy (similar to faithfulness)
and fluency (related to grammaticality), has been used in machine translation (MT)
evaluation \cite[e.g.,][]{banchs2015adequacy}.	
%Specifically, meaning preservation RLM is also cross-lingual and may prevent meaning changing output that is over-fitted to other measures (such as one word output to get high precision, or always outputting the same sentence to receive full grammaticality score).\lc{Perhaps too many guesses and assumptions?} 

As a test case, we use the UCCA semantic scheme \cite{abend2013universal},
motivated by its recent use in semantic evaluation of MT \cite{birch2016hume} and text simplification \cite{sulem2018Semantic} systems.
Nevertheless, {\sc USim} can be easily adapted to other semantic schemes, such as AMR \cite{banarescu-EtAl:2013:LAW7-ID}.
{\sc USim} is conceptually related to RLMs developed
for MT
\cite{reeder2006measuring,albrecht2007regression,specia2009estimating,specia2010machine}.
Notably, XMEANT \cite{lo2014xmeant} compares the source to the output
in terms of their semantic role labeling structures.
Our use of UCCA is motivated by its wider coverage of predicate types, as opposed to 
MEANT's focus on verbal predicates, and UCCA's preservation
of structure across translations \cite{sulem2015conceptual}.
See \cite{birch2016hume} for further discussion. 

%{\sc USim} allows changes to the source, but only to the extent that these changes
%do not alter the semantic structure. 

%We show that due to the vast number of possible valid corrections, reference-based measures suffer from 
%biases that will not be solved by more references. 
%Specifically, systems developed towards such measures 
%hardly correct, state-of-the-art systems are such systems. 

%Semantic annotation schemes are also gaining support and proven to be useful for many tasks in general\lc{have preference on what to cite?} and specifically for evaluation \cite{birch2016hume}. One desired feature of semantic annotation is its robustness to modifications that do not change the meaning such as paraphrasing \cite{abend2013universal}, and translation \cite{sulem2015conceptual}.

We conduct experiments to confirm {\sc USim}'s validity.
Specifically, we show that
(1) UCCA can be consistently and automatically applied to learner language (LL) (\S \ref{subsec:valid_faithfulness}), 
(2) {\sc USim} is not prone to unduly penalize valid corrections (\S \ref{subsec:valid_faithfulness}), and 
(3) {\sc USim} assigns a lower score to corrections of poor quality (\S \ref{subsec:sensitivity}).
Our experiments also indicate that UCCA parsing technology is already sufficiently mature for an automatic variant of {\sc USim} to provide reliable results (\S \ref{subsec:automatic}).

\section{Background}

\paragraph{LL Annotation.}
While most linguistic theories propose that each learner makes consistent use of syntax 
\cite{huebner1985system,tarone1983variability}, this use may not conform to the syntax of the learned language, 
or of any other known language. This entails difficulties in defining syntactic annotation for LL, as the annotated syntax differs between learners.

Syntactic schemes for LL annotate syntactically erroneous sentences in different ways.
\newcite{berzak2016universal} and \newcite{ragheb2012defining} annotate according 
to the syntax used by the learner, even if this use is not grammatical.
Such annotation may be unreliable for measuring faithfulness, 
as GEC systems aim to alter these erroneous syntactic structures.
\newcite{nagataphrase} take the opposite approach, and remain faithful to the syntax intended by the learner.
This has also been the tradition in works on parser robustness 
\cite{bigert2005unsupervised,foster2004parsing}. However, such approach is prone to inconsistencies due to the 
variety of different syntactic structures that can be used to express a similar meaning. 

In this paper, we use semantic annotation to structurally
represent LL. Semantic structures are faithful to the intended
meaning, and not to the formal realization, and thus face
fewer conflicts where the syntactic structure used diverges from
the one intended. We are not aware of any previous attempts to semantically
annotate LL text.

\paragraph{The UCCA Scheme.}\label{sec:ucca}
UCCA is a semantic annotation scheme that builds on
typological and cognitive linguistic theories.
The scheme's aims are to provide a coarse-grained, cross-linguistically
applicable representation.
Importantly, UCCA's categories directly reflect semantic, rather than
distributional distinctions.
For instance, UCCA is not sensitive to POS distinctions:
a Scene's main relation can be a verb but also an adjective
(``He is {\bf thin}'') or a noun (``John's {\bf decision}'').
Indeed, \newcite{sulem2015conceptual} have found that UCCA structures are
preserved remarkably well across English-French translations. 

UCCA structures are directed acyclic graphs, where the words
correspond to (a subset of) their leaves.
The nodes of the graphs, called {\it units}, are either leaves or several elements jointly
viewed as a single entity according to some semantic or cognitive consideration.
The edges bear one or more categories, indicating the role of 
the sub-unit in the relation that the parent represents.

UCCA views the text as a collection of {\it Scenes} and relations between them.
A Scene describes a movement, 
an action or a state which is persistent in time.
Every Scene contains one main relation, 
zero or more {\it Participants}, 
interpreted in a broad sense to include locations, destinations and complement clauses,
and {\it Adverbials}, such as manner or aspectual modifiers.

\com{
\begin{figure}[t]
				\begin{tikzpicture}[sibling distance=10mm, level distance=10mm, ->,
				every node/.append style={midway},
				every circle node/.append style={fill=black}]
				{
					\node (Source) [circle] {}
					child {node (He) {He} edge from parent node[left] {\scriptsize A}
					}
					child {node (gave) {gve} edge from parent node[right] {\scriptsize A}}
					child {node (an apple) [circle] {}
						{
							child {node (an) {an} edge from parent node[left] {\scriptsize E}}
							child {node (apple){apple} edge from parent node[right] {\scriptsize C}}
						} edge from parent node[right] {\scriptsize P} }
					;}
					\end{tikzpicture}

		\end{figure}}

\section{Semantic Faithfulness Measures}\label{sec:Semantics}
\begin{figure}[t]
		\hspace{-1.8cm}
		\begin{subfigure}{1.1\columnwidth}

			\parbox{.8\columnwidth}{\hspace{1.5cm}
				\scalebox{.9}{
					\begin{tikzpicture}[sibling distance=1mm, level distance=7mm,
					every node/.append style={midway},
					every circle node/.append style={fill=black}]
					\begin{scope}[frontier/.style={distance from root=15mm}, ->]
					\Tree [.\node [circle] (rootu) {};
					\edge node [auto=right]{\scriptsize A,1}; \node (Heu) {He};
					\edge node[auto=right down]{\scriptsize P,2}; \node (gve) {gve};
					\edge node[auto=right]{\scriptsize A,4};
					[.\node [circle](an appleu) {};
						\edge node[auto=right]{\scriptsize E,5}; \node (anu) {an};
						\edge node[auto=left]{\scriptsize C,6}; \node (appleu) {apple};
					]
					\edge node[auto=left]{\scriptsize A,3};
					[.\node [circle](for john) {};
					\edge node[auto=right]{\scriptsize R,Empty};\node (for) {for};
					\edge node[auto=left]{\scriptsize C,3}; \node (john) {john};
					]]
					\end{scope}
					\begin{scope}[yshift=-3.5cm,grow'=up, frontier/.style={distance from root=12mm}, ->]
					\Tree [.\node [circle] (rootd) {};
					\edge node [auto=left]{\scriptsize A,1}; \node (Hed) {He};
					\edge node[auto=right]{\scriptsize P,2}; \node (gave) {gave};
					\edge node[auto=right]{\scriptsize A,3};\node (John) {John};
					\edge node[auto=right]{\scriptsize A,4};
					[.\node [circle] (an appled) {};
					\edge node[auto=left]{\scriptsize E,5}; \node (and) {an};
					\edge node[auto=right]{\scriptsize C,6}; \node (appled) {apple};
					]
]
					\end{scope}
					\begin{scope}[dashed]
					\draw (Heu) -- (Hed);
					\draw (gve) -- (gave);
					\draw (John) -- (john);
					\draw (anu) -- (and);
					\draw (appleu) -- (appled);
					\end{scope}
					\end{tikzpicture}
				}}
				\parbox{.1\columnwidth}{
					\begin{adjustbox}{width=.27\columnwidth,margin=1pt,frame}
						\begin{tabular}{ll}
							P & process \\
							A & participant \\
							H & linked scene \\
							R & relator \\
							C & center \\
							E & elaborator \\
						\end{tabular}
					\end{adjustbox}
				}
			\end{subfigure}

				\caption{\label{fig:example}
					UCCA structures of a learner language (top) and correction (bottom) including word alignments (dashed). On the edges are labels and numbers aligned to (top) or indexes (bottom). Precision is $\frac{7}{9}$ Recall is $\frac{7}{7}$.
				}

			\end{figure}
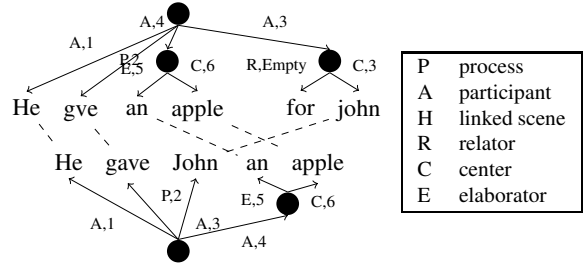
We start by defining a simplified measure, used for inter-annotator agreement (IAA).
The measure compares two UCCA annotations over the 
same set of tokens. We then proceed to define {\sc USim}, which compares two UCCA structures
over alignable but different sets of tokens.

\paragraph{IAA Measure.} We define a similarity measure over UCCA annotations 
$G_1$ and $G_2$ that share their set of leaves (tokens) $W$.
For a node $v$ in $G_1$ or $G_2$, define its yield $yield(v) \subseteq W$ as its
set of leaf descendants.
Define a pair of edges $(v_1,u_1) \in G_1$ and $(v_2,u_2) \in G_2$ to be matching
if $yield(u_1) = yield(u_2)$ and they have the same label.
Labeled precision and recall are defined by dividing the number of matching edges
in $G_1$ and $G_2$ by $|E_1|$ and $|E_2|$ respectively.
{\it DAG $F$-score} is their harmonic mean.
The measure collapses to the common parsing $F$-score if $G_1, G_2$ are trees.

\paragraph{The {\sc USim} Measure.} Computing a faithfulness
measure is slightly more involved, as the source sentence graph $G_s$ and its
correction $G_c$ do not share the same set of leaves.
%Giving a more accurate
%measure than the upper bound suggested by \cite{sulem2015conceptual} for
%comparing two parallel texts in different languages, while keeping
%the essence of comparing how many of the aligned nodes conserve meaning and tag. For that we may think for a moment on GEC as
%translation from LL to Proper English, and a good translation
%would be a translation which keeps the meaning but has the syntax
%of English. Considering that, just like in translation, we can align words from the LL to the corresponding words in English
%and keep record of how many of those nodes kept their labels.
%
%As comparing labels is trivial, and done before us. We should focus on how we propose to align nodes. 
%First, note that comparison should not be at
%the token level, as we want to allow tokens to be corrected - replaced or removed -
%as long as the higher structures convey the same meaning. We thus
%prune the s above the leaves, the tokens of the sentence. To
%define an alignment of the nodes, we suggest some possible ways, all
%based on first aligning the words in order to give order to the DAG and then comparing the structure in one way or another.
%
We assume a (possibly partial, possibly many-to-1) alignment between $G_s$ and $G_c$,
$A \subset V_s \times V_c$. 

An edge $(v_1,v_2) \in E_c$ is said to match an edge
$(u_1,u_2) \in E_s$ if they have the same label and $(v_2,u_2) \in A$. Recall (Precision)
is defined as the ratio of edges in $E_s$ ($E_c$) that have a match in $E_c$ ($E_s$) respectively, and
$F$-score is their harmonic mean. We note that this measure collapses to the
{\it DAG $F$-score} if $A$ includes all pairs of nodes in $E_s$ and $E_c$ that have
the same yield. See Figure \ref{fig:example}. 

In order to define the alignment between $V_s$ and $V_c$, we begin by aligning the leaves
(tokens) in $V_s$ and $V_c$.
Alignment is cast as a weighted bipartite graph matching problem. Edge weights are assigned to be the edit distances between the tokens.
We note that aligning words in GEC (and other monolingual translation tasks) is much simpler than in MT,
as most of the words are unchanged, deleted fully, added, or changed slightly.
Denote the resulting leaf alignment with $A_l \subset Leaves_s \times Leaves_c$.
We extend $A_l$ to define the node alignment $A$, aligning each non-leaf $v \in V_s$
to the node $u \in V_c$ that maximizes

\begin{small}
 \begin{myequation*}
  w\left(v,u\right) = \frac{\vert A_l \cap \left(yield\left(u\right) \times yield\left(v\right)\right)\vert}{\vert yield\left(u\right) \vert}.
 \end{myequation*}

\end{small}

\noindent
We exclude from $A$ zero-weighted pairs.
{\sc USim} is defined to be the $F$-score resulting from $A$.
As the alignment may differ when aligning nodes from $V_c$ to $V_s$
and the other way around, we report {\sc USim} in both directions.

{\sc USim} is somewhat more relaxed than {\it DAG $F$-score},
as, unlike DAG $F$-score, it also aligns nodes whose yields are not in perfect alignment with one another.
This relaxation is necessary, given that corrections often add
or remove nodes, thus eliminating the possibility of a perfect alignment.
In order to obtain comparable IAA scores, we report IAA using {\sc USim} as well.
%
%Then for each node in $v \in V_s$, we compute its descendent leaves $yield(v) \subset Leaves_s$ as before,
%and compute their projection $yield'(v) = \{u \in Leaves_c:(v,u) \in A_l\}$.
%We now define the node alignment to be $A = \{(u,v) \in V_s \times V_c : yield'(u) = yield(v) \}$.
%We note that $A_l \subset A$ and that if $V_s$ and $V_c$ share the same tokens,
%this computation again reduces to DAG $F$-score.

For completeness, we replicate the protocol used by \newcite{sulem2015conceptual}
for comparing the UCCA annotations of standard English-French translations, which we call
Distributional Similarity ({\sc DistSim}).
For a given UCCA label $l$, $c_i(l)$ is the number of $l$-labeled UCCA edges
in the i-th source sentence, and $d_i(l)$ is the number of $l$-labeled UCCA edges
in its corresponding correction. We define {\sc DistSim}(l) between these
sentences to be $\frac{1}{N}\sum_{i=1}^N \vert c_i(l) - d_i(l) \vert$, where
$N$ is the total number of sentence pairs.

\section{Experiments}

We conduct four types of experiments to validate {\sc USim}, showing that:
(1) semantic annotation can be consistently applied to LL through inter-annotator agreement (IAA) experiments;
(2) a valid corrector scores high on {\sc USim};
(3) an automatic UCCA parser can reliably replace human annotation for {\sc USim};
(4) {\sc USim} is sensitive to changes in meaning.
%by comparing the semantic structures of the source to corrections of fairly poor quality.

%\vspace{-.2cm}
\subsection{Experimental Setup.}
We train two UCCA annotators, the first author and a paid in-house annotator by annotating both LL and standard English
passages, until a high enough agreement is reached (6 training hours).
Training passages are excluded from the evaluation.
We use UCCA's annotation
guidelines\footnote{\url{http://www.cs.huji.ac.il/~oabend/ucca.html}}
without any adaptations.

We experiment on 7 essays and their corrections, each comprising about 500 tokens (see supplementary material \ref{tab:annotated-paragraphs}).
In order to measure IAA, we assigned 4 of these essays to both annotators.
In order to measure the faithfulness score for a valid correction,
we annotate both the source
and the manually corrected versions of 6 essays,
3 of which were annotated by both annotators.

\subsection{The Faithfulness of Valid Corrections.}\label{subsec:valid_faithfulness}
We obtain an IAA {\it DAG $F$-score} of 0.845
(Precision 0.834, Recall 0.857), which
is comparable to the IAA reported for English Wikipedia texts by \citet{abend2013universal}.
As another point of comparison, we doubly annotate 3 corrected
NUCLE \cite{dahlmeier2013building} passages, obtaining a similar IAA.
These results suggest that UCCA annotating LL does not degrade IAA:
it can be applied as consistently to LL as to standard English.

Table \ref{tab:Distances} (left-hand side) presents the {\sc USim} scores obtained by comparing 
the NUCLE references and the source, or equivalently the score of a valid correction.
To control for differences between the annotators, we explore both
a setting where both sides are annotated by the same annotator,
and a setting where they are annotated by different ones.
As an upper bound on the score of a valid corrector (using different annotators),
we also report the {\sc USim} IAA on source sentences. 

Our results indicate that a valid correction obtains a score comparable
to the IAA, which indicates that {\sc USim} is indeed
insensitive to the surface divergence between a source sentence and its valid corrections.
Finally, we compute the {\sc DistSim} measure
between the source and reference sentences (Table \ref{tab:Distances}, right-hand side),
obtaining similar results to those obtained by \newcite{sulem2015conceptual}. It suggests that on a coarse grained level, UCCA structures are as robust to grammatical error corrections as they are to translation from English to French, which was shown to be very robust, specifically more robust than syntactic representation \cite{sulem2015conceptual}.
\begin{table}
	%\vspace{-0.5cm}
	\small
	\centering
	\singlespacing
	\begin{tabular}{c|c|c|c||c|c|}
		\cline{2-6} 
		& \multicolumn{3}{c||}{\sc USim} & \multicolumn{2}{c|}{\sc DistSim}\\ \cline{2-6}
		& s$\rightarrow$r & r$\rightarrow$s & Avg & A+D & Scene\
		\\
		\hline
		Different & 0.85 & 0.83 & 0.84 & 0.96 & 0.93
		\\
		Same & 0.92 & 0.91 & 0.92 & 0.97 & 0.96
		\\
		\hline
		\hline
		IAA & 0.85 & 0.81 & 0.83 & - & -
		\\
		\hline
		SAR15 & - & - & - & 0.95 & 0.96 \\
		\hline
	\end{tabular}
	\caption{\label{tab:Distances}
		The faithfulness of valid corrections.
		The left-hand side presents {\sc USim},
		where s$\rightarrow$r is the setting where alignment is computed from the source to the reference,
		r$\rightarrow$s is the other way around, and Avg is their average.
		The right-hand side presents {\sc DistSim} for the UCCA categories Participants and Adverbials
		together (A+D), and for Scenes (Scene).
		Rows indicate whether the same annotator annotated the source and reference or not.
		For comparison, the IAA row is the IAA computed using {\sc USim}.
		Results show that the valid corrector's faithfulness is comparable with IAA.
		SAR15 are reported by Sulem et al. on English-French
		translations; similarity is comparable to ours.}
	%\vspace{-0.6cm}
\end{table}

\subsection{Automatic {\sc USim}.}\label{subsec:automatic}

We experiment with an automatic variant of {\sc USim}, where UCCA
structures are parsed automatically.
We use the TUPA parser \cite{hershcovich2017transition} to generate UCCA structures,
instead of the human annotators. Otherwise the setup is as above.
TUPA is used with its biLSTM model, trained on the UCCA English Wikipedia corpus.

We obtain a {\sc USim} score of 0.7 between the parses of the reference
correction and the source, which is comparable to the parser's reported
performance (0.73 in-domain, 0.68 out-of-domain), despite not performing any
domain adaptation to LL. 
That is, the UCCA parses of the source and the correction are roughly as similar to each
other as they are to their gold standard parse. This supports the hypothesis 
%\lc{Originally it was written indicates, a thorough reviewer pointed out that had we not received the results it would indicate parsing is not mature, given our experiment it is more likely to be mature but it is not proving the claim.} 
that semantic parsing technology is sufficiently mature to
be applicable to {\sc USim}.
Results also suggest an improvement in parsing performance may further improve these scores.

%We note that the parser was not trained again in order to capture LL. Still, the amount of agreement is more or less the score the parser gets.

\com{
	\begin{table}
		\centering
		\singlespacing
		\begin{tabular}{c|c|c|c|}
			\cline{2-4} 
			& \multicolumn{3}{c|}{\sc USim} \\
			\cline{2-4}
			& s$\rightarrow$r & r$\rightarrow$s & Avg\
			\\
			\hline
			TUPA & 0.7 & 0.7 & 0.7
			\\
			\hline
			\hline
			Different & 0.85 & 0.83 & 0.84
			\\
			\hline
		\end{tabular}
		\caption{\label{tab:parser} The table presents {\sc USim}
			where the alignment is computed from the source to the reference (s$\rightarrow$r),
			the opposite direction (r$\rightarrow$s), and their average (Avg).
			The first row presents results using TUPA parser \cite{hershcovich2017transition}.
			The second row we see the results of one annotator for the source and another for the reference.
			The results show that the valid corrector's faithfulness is captured quite
			well with the automatic parsing, around the parser reported accuracy and standard English.}
	\end{table}
}

\subsection{Sensitivity to Error Types}\label{subsec:types}
To provide another perspective on automated {\sc USim}'s behaviour,
we examined the measure's sensitivity to different error types, using {\sc MAEGE} \cite{choshen2018maege}. 
For each NUCLE sentence and set of edits (replacements of sub-strings that contain an error by corrected ones. Such edit for the example in fig. \ref{fig:example} might be $"gva"\rightarrow"gave"$, with type $spelling$), we sample an order in which edits are applied.
We select the source randomly to be one of the resulting sentences. 
We then compare the difference in {\sc USim} before and after applying each edit, and average these differences by the applied edit type. 
We denote the average difference in {\sc USim} due to correction of errors of type $t$ with $\Delta_t$.
The hypothesis is that $\Delta_t$ should be close to 0 for all $t$, as edits are manual and are thus assumed to be faithful.
We focus on edit types with high $|\Delta_t|$ to better understand where {\sc USim} fails. See table \ref{tab:MAEGE} in the supplementary material for complete results.

We find that among the 5 most penalized error types by {\sc USim} are ``unclear meaning'' and corrections of type ``other'', that fit no specific type; these corrections may actually change the meaning of the original sentence. In the most penalized and most rewarded changes we see ''Dangling Modifier``, ''Pronoun Reference`` and ''Word Tone`` errors, the first usually changes a word into a more complex structure and the latter two the opposite. Such changes alter the lower levels of the UCCA structure (near the leaves); a similarity measure that focuses on the top of the DAG, or one that performs a better lexical semantic abstraction, may address this sensitivity. Corrections of incorrect word order are also highly rewarded (high $\Delta_t$), probably due to parser performance (the UCCA structures themselves are not affected by word order). Training the parser with LL annotated data may address this sensitivity.
	
Among the most rewarded changes we also see errors of replacing rare or misconstructed words with proper English words (Acronym and Mechanical errors). We assume this is due to parser performance, as TUPA only extracts features over complete words, and has no character-level encoding at this point. Thus, all misconstructed words fall into an out-of-vocabulary category and can only be labeled by the context.

Lastly, adding a missing verb is shown to be highly rewarded. Under the UCCA guidelines, a missing verb should be annotated as an implicit unit, but as TUPA does not generate implicit units, it is not surprising that when corrections transforms an implicit unit into an explicit word, the parser's output changes (and hence {\sc USim}). Future improvements to TUPA may address this.

%\vspace{-.2cm}
\subsection{Sensitivity to Unfaithfulness.}\label{subsec:sensitivity}

We have shown that UCCA is insensitive to differences between a source sentence
and its valid correction. We now present an evaluation of the sensitivity of {\sc USim}
to proposed corrections that diverge semantically from the source.
%To wrap up the argument we show that {\sc USim} is not only insensitive to corrections (Table \ref{tab:Distances}) but also sensitive to meaning changing corrections.
A semantic measure is, by its definition, sensitive to variation in
the semantic dimensions which it encodes. 
In UCCA's case, these distinctions focus on predicate-argument structures,
the inter-relations between them, and the semantic heads of complex arguments.
These distinctions are widely regarded as fundamental in the NLP and linguistic literature.

In order to empirically validate this claim, we present an experiment which shows that corrections
of a fairly low quality indeed receive a much lower {\sc USim} faithfulness score.
Current state-of-the-art systems rarely alter the source sentences enough to yield semantically unfaithful outputs \cite{choshen2018conservatism}.
Consequently, their human rankings are not determined by their semantic faithfulness, rendering them unuseful for validating {\sc USim}.
We instead experiment with 5 partially trained correctors, trained and evaluated on the
JFLEG corpus \cite{napoles2017jfleg} by \newcite{sakaguchi2017grammatical}.

{\sc USim} is computed automatically for each system's output on 754 source sentences.
Low faithfulness results are expected, as these outputs include major changes,
sometimes deleting full phrases from the output or changing every other word.
Indeed, automatic {\sc USim} obtains scores of 0.32-0.39 for 4 of the systems, and 0.19
for the system that obtains the lowest GLEU \cite{napoles2015ground} score.
For completeness, we run {\sc USim} on the 4 references provided by JFLEG for each
source and obtain scores of 0.72-0.78, suggesting the domain change is not the reason for the low {\sc USim} score.

Taken together, these results indicate that {\sc USim}, even in its automatic variant,
is sensitive to semantic changes. Consider the example: 

\begin{table}[h!]
	%\vspace{-.3cm}
  \centering
  \label{ex:sensitive}
  \begin{tabular}{p{0.2\columnwidth}p{0.73\columnwidth}}
    Source    & \small the good student must know how to understand and work hard to get the iede.\\
    Reference & \small A good student must be able to understand and work hard to get the idea.\\
    Corrector & \small The good student must know how to understand and work hard to get on.     
  \end{tabular}
  
  %\vspace{-.3cm}
\end{table}

{\sc USim} assigns the reference 0.71 and only 0.33 to the corrector.
Moreover, although the reference makes more word changes than the proposed correction,
it still obtains a higher {\sc USim} score.

%%%%%%%%%%%%%%%%%%%%%%%%%%%%%%%%%%%%%%%%%%%%%%%%%%%%%%%%%%%%%%%%%%%%
%
%was shown to have some wanted attributes that the $M^2$ scorer lacks.
%For example, it scales from -1 to 1 providing a way to know if a correction is an improvement over the
%source. I-measure expects the same input as $M^2$ but its score differs as it is based on token-level
%edits accuracy score. 
%
%Addressing the need to improve automatic GEC evaluation, three sophisticated measures have been proposed,
%all of them are reference based. 
%$M^2$ was introduced for CoNLL2013, providing a way to compute phrase-level edits $F$-Score. As an
%input $M^2$ expects a source sentence, a correction and a set of edits for each reference in the
%gold standard.
%It uses an edit lattice to optimistically choose edits for the reference that
%will best match those of the references. Since it was introduced $M^2$ is the standard scorer for GEC.
%
%%%%%%%%%%%%%%%%%%%%%%%%%%%%%%%%%%%%%%%%%%%%%%%%%%%%%%%%%%%%%%%%%%%%

\section{Conclusion}

We propose a measure of semantic faithfulness of a correction to the source,
thereby avoiding the pitfalls of reference-based evaluation. 
We believe that using RLMs in conjunction with RBMs in the training and development of GEC
systems will better address the challenge of over-conservatism, and the 
high costs of acquiring many references.

Future work will conduct user studies to assess the relative importance
of different evaluation criteria.
Specifically, we will explore to what extent users are
tolerant to invalid changes to the sentence's structure, i.e.,
violation of conservatism, relative to their tolerance to invalid changes 
to the sentence's meaning, i.e., violation of faithfulness.
A better understanding of how these interact
may lead to improved semantic evaluation that will alleviate the need
for a high number of references.

\section*{Acknowledgments}

This work was partially supported by the Israel Science Foundation
(grant No. 929/17), as well as by the HUJI Cyber Security Research
Center in conjunction with the Israel National Cyber Bureau in the
Prime Minister's Office.
We thank Joel Tetreault and Courtney Napoles for helpful feedback and help with obtaining the partially trained GEC system outputs.

\bibliographystyle{acl_natbib}
\bibliography{propose}

\end{document}

\appendix

\title{}
%\maketitle

\onecolumn
\section*{\centering\Large Supplementary Material for ``Faithfulness reference-less measure for grammatical error correction''}
\section{Annotated paragraphs}
	\begin{table}[hb]
		\centering
		\begin{tabular}{lll}
			Annotator-id & NUCLE-id & type      \\
			1         & 2  & corrected \\
			2         & 2  & corrected \\
			1         & 2  & learner   \\
			2         & 2  & learner   \\
			1         & 3  & corrected \\
			2         & 3  & corrected \\
			1         & 3  & learner   \\
			2         & 3  & learner   \\
			1         & 5  & corrected \\
			2         & 5  & corrected \\
			1         & 5  & learner   \\
			2         & 5  & learner   \\
			1         & 6  & learner   \\
			2         & 6  & learner   \\
			2         & 7  & corrected \\
			2         & 7  & learner   \\
			1         & 8  & corrected \\
			1         & 8  & learner   \\
			1         & 10 & corrected \\
			1         & 10 & learner  
		\end{tabular}
		\caption{The list of paragraphs annotated, showing which annotator annotated it, which type of language is used in it and the corresponding id in the NUCLE corpus. Note that parallel paragraphs have the same id.\label{tab:annotated-paragraphs}}
	\end{table}

\npdecimalsign{.}
\nprounddigits{3}
\begin{table}[]
	\centering
		\begin{tabular}{@{}ln{5}{2}l@{}}
			\toprule
			& \multicolumn{1}{l}{Average change}  & \multicolumn{1}{l}{Occurrences} \\ \midrule
			Wtone    & 0.1861439688    & 28         \\
			Mec      & 0.07307186292   & 731        \\
			Wa       & 0.07235489879   & 2          \\
			V0       & 0.06299936463   & 74         \\
			Pref     & 0.05599492536   & 235        \\
			Ssub     & 0.05578719858   & 89         \\
			WOadv    & 0.02061675473   & 30         \\
			Sfrag    & 0.01952410882   & 15         \\
			Rloc-    & 0.01841189615   & 336        \\
			ArtOrDet & 0.01672618109   & 893        \\
			Vm       & 0.01642930822   & 92         \\
			SVA      & 0.01565099167   & 270        \\
			Wci      & -0.001632237447 & 869        \\
			Trans    & -0.004670224979 & 192        \\
			Srun     & -0.004856333151 & 30         \\
			Vt       & -0.005917533634 & 286        \\
			Prep     & -0.01062070433  & 634        \\
			Pform    & -0.01241762832  & 76         \\
			Vform    & -0.01523024009  & 250        \\
			Npos     & -0.02472584516  & 54         \\
			Wform    & -0.02806091938  & 192        \\
			Nn       & -0.02863355141  & 479        \\
			Um       & -0.03762075133  & 61         \\
			Others   & -0.05341553854  & 93         \\
			WOinc    & -0.08941563292  & 80         \\
			Smod     & -0.0894551682   & 6          \\
			Spar     & -0.1006425275   & 30         \\ \bottomrule
		\end{tabular}%
	
	\caption{{\sc USim} average change in score for applying human corrections by edit types (abbreviations in \citet{dahlmeier2013building}). For completeness, the number of corrections considered from each type is reported in the rightmost column.\label{tab:MAEGE}}
	
\end{table}
\FloatBarrier

\bibliographystyle{acl_natbib}
\bibliography{propose}